\documentclass[twocolumn]{cup-pan}
\usepackage[utf8]{inputenc}
\usepackage{blindtext}
\usepackage{dblfloatfix}
\usepackage[english]{babel}
\usepackage[numbers]{natbib}
\usepackage{multicol}
\usepackage{float}
\usepackage{geometry}
\usepackage{hyperref}
\hypersetup{
    colorlinks=true,
    linkcolor=blue,
    filecolor=magenta,
    urlcolor=cyan,
}
\title{Deep learning methods for screening patients' S-ICD implantation eligibility}

\author{{\small{Anthony J. Dunn$^{a}$, Mohamed H. ElRefai$^{b}$, Paul R. Roberts$^{b}$, Stefano Coniglio$^{a}$, Benedict~M.~Wiles$^{c}$ and Alain~B.~Zemkoho$^{a}$}}\\
\emph{\small{$^{a}$ School of Mathematical Sciences, University of Southampton, United Kingdom}}\\
\small{$\{$\textit{ajd1g15}, \textit{a.b.zemkoho}, \textit{s.coniglio}$\}$\textit{@soton.ac.uk}}\\
\emph{\small{$^{b}$ University Hospital of Southampton, United Kingdom}}\\
\small{$\{$\textit{mohammed.elrefai}, \textit{paul.roberts}$\}$\textit{@uhs.nhs.uk}}\\
\emph{\small{$^{c}$ St George’s University Hospitals NHS Foundation Trust, United Kingdom}}\\
\small{$\textit{benedict.wiles@gmail.com}$}}
\begin{document}
\twocolumn[
    \begin{@twocolumnfalse}
        \maketitle

        \begin{abstract}

        {

        Subcutaneous Implantable Cardioverter-Defibrillators (S-ICDs) are used for prevention of sudden cardiac death triggered by ventricular arrhythmias. T Wave Over Sensing (TWOS) is an inherent risk with S-ICDs which can lead to inappropriate shocks. A major predictor of TWOS is a high T:R ratio (the ratio between the amplitudes of the T and R waves). Currently patients' Electrocardiograms (ECGs) are screened over 10 seconds to measure the T:R ratio, determining the patients' eligibility for S-ICD implantation. Due to temporal variations in the T:R ratio, 10 seconds is not long enough to reliably determine the normal values of a patient's T:R ratio. In this paper, we develop a convolutional neural network (CNN) based model utilising phase space reconstruction matrices to predict T:R ratios from 10-second ECG segments without explicitly locating the R or T waves, thus avoiding the issue of TWOS. This tool can be used to automatically screen patients over a much longer period and provide an in-depth description of the behaviour of the T:R ratio over that period. The tool can also enable much more reliable and descriptive screenings to better assess patients' eligibility for S-ICD implantation.
        }
        \end{abstract}
     \end{@twocolumnfalse}
]

\section{Introduction}\label{Introduction}
Sudden Cardiac Death (SCD) is one of the leading causes of death in the modern world. Most deaths caused by SCD can be attributed to Ventricular Arrhythmias (VA)~\cite{adabag2010sudden}. The key to survival in patients who are affected by VA is adequate Cardiopulmonary Resuscitation (CPR) and early defibrillation~\cite{hazinski2005lay}. Current medical guidelines recommend the use of Implantable Cardioverter-Defibrillators (ICDs) for prevention of SCD triggered by VA in high risk populations~\cite{kusumoto2018systematic, authors20152015}. Conventional transvenous ICDs (TV-ICDs) consist of a \textit{can} and transvenous leads implanted into the right ventricle to treat the arrhythmia by delivering a voltage shock. TV-ICDs are associated with the risk of complications with potentially fatal consequences.
\begin{figure}[!b]
  \centering
  \includegraphics[width=\columnwidth]{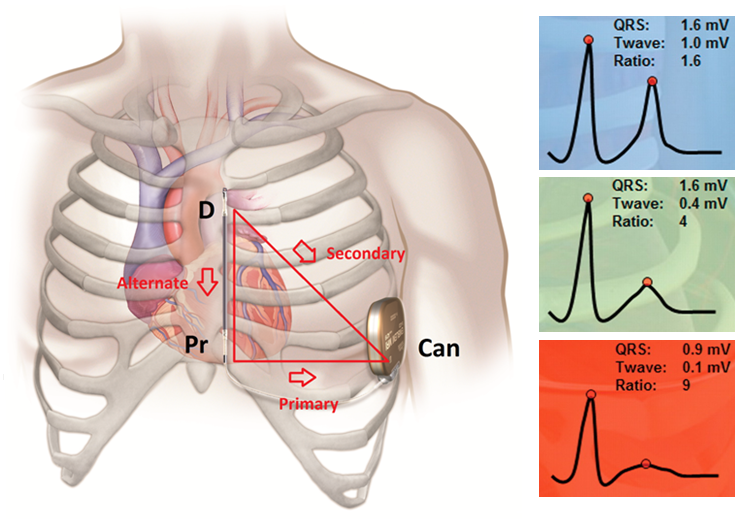}
  \caption{S-ICD sensing electrodes and vectors between them. An implanted S-ICD with underlying anatomical features showing the location of the can (pulse generator), the proximal (Pr) and distal (D) sensing electrodes and the shocking coil (located between the electrodes) Image (prior to annotation) © Boston Scientific Corporation or its affiliates. Reproduced with permission.}\label{SICD}
\end{figure}

The Subcutaneous ICD (S-ICD), which comprises an electrically active can and a single subcutaneous lead (see Figure \ref{SICD}) was designed to avoid complications of the TV-ICD by utilising a totally avascular approach. The sensing mechanism of the S-ICD has been shown to be equally effective to that of the
\begin{figure*}[!b]
  \centering
  \includegraphics[width=14cm]{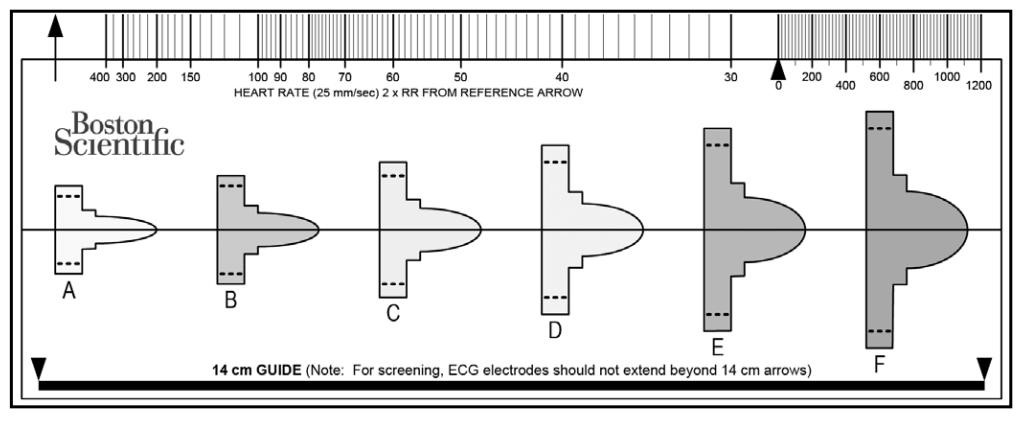}
  \caption{S-ICD screening tool. The recorded QRST morphology in every vector is then compared to the acceptable templates. The template is aligned to the isoelectric line of the ECG, and the QRST complexes are viewed through the appropriately sized template. The R wave peak of the ECG must be placed within either hashed box (positive or negative) of any template. A vector passes screening if the remainder of the QRST complex sits entirely within the boundary of the template. (this manual screening method is now largely replaced by the manufacturer with an automatic screening tool following the same principals)}\label{ScreeningTool}
\end{figure*}TV-ICD~\cite{boersma2017implant} and demonstrated less incidence of device related complications when compared with conventional ICDs~\cite{knops2020subcutaneous}. However, a consequence of the algorithm used by the S-ICD to detect VA is an inherent risk of T Wave Over Sensing (TWOS), whereby the T wave, one of the 5 main waveforms of the PQRST complex (the electrocardiogram (ECG) of a single heartbeat), which follows a QRS complex (comprised of the Q, R and S waves of a PQRST complex) is misinterpreted as a second R wave, which can lead to inappropriate shocks. Inappropriate shocks are associated with increased morbidity and mortality~\cite{van2011inappropriate}.

\begin{figure}[!t]\label{HolterPositions}
  \centering
  \includegraphics[width=\columnwidth]{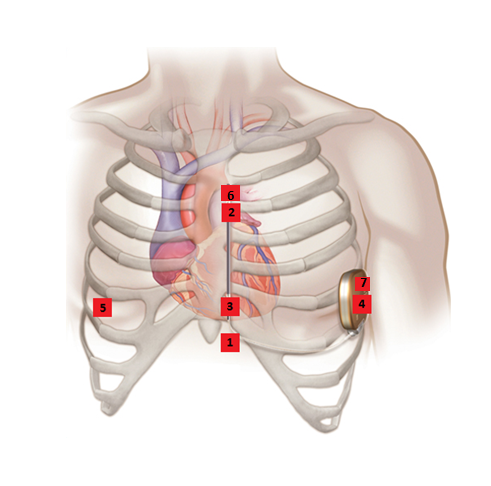}
  \caption{Holter recorder surface ECG positions. }\label{HolterPositions}
\end{figure}
Not all patients are eligible for S-ICD therapy and eligibility is identified during a pre-implant screening process that is undertaken in potential S-ICD recipients. Surface ECGs are used as a surrogate marker of future S-ICD vectors (as shown in Figure \ref{SICD})to non-invasively assess S-ICD eligibility. These ECG recordings are assessed. Vectors with lower T:R ratios (the ratio between the amplitude of the T wave and that of the R wave) are more likely to pass the screening, while patients with an ECG morphology that does not meet this screening criteria are deemed at high risk of TWOS and are ineligible for an S-ICD. Despite the current screening practice, the most common cause of inappropriate shocks in patients implated with S-ICDs remains TWOS~\cite{knops2020subcutaneous}. The T:R ratio---a major predictor of S-ICD eligibility---is not fixed in the same individual because of frequently observed temporal variations in the amplitudes of the R and T waves which are influenced by multiple factors~\cite{madias2001anasarca,madias2005qtc,fosbol2008prognostic,assanelli2013t}. In patients with S-ICDs, variations in the T:R ratio often go undetected as “silent” episodes of TWOS in spite of carrying a considerable risk of leading to the development of clinically relevant oversensing, which manifests in inappropriate shocks.

Under the current screening process, ECG electrodes are placed on the chest wall using the anatomical landmarks which correspond to the location of the sensing electrodes of the potential future S-ICD implantation shown in Figure \ref{HolterPositions}. A short three lead ECG recording (several PQRST complexes), corresponding to the three vectors utilised by the S-ICD, is evaluated using the manual S-ICD screening tool shown in Figure \ref{ScreeningTool}. Henceforth, when discussing leads, we refer only to the three leads corresponding to the three vectors used by the S-ICD.
The patient is deemed eligible for S-ICD implantation if, for at least one of the leads (representing one of the S-ICD vectors), the patient's QRST complex (the PQRST complex excluding the P wave) sits entirely within the boundary of the template. These templates correspond roughly to a maximum acceptable T:R ratio of 1:3. As mentioned previously, the T:R ratio could fluctuate, as the amplitudes of both R waves and T waves may vary according to other factors (e.g, electrolyte levels). Due to the short duration (several PQRST complexes) of the current screening, it is possible that a patient with a typically high T:R ratio could pass this screening and likewise a patient with a typically low T:R ratio could fail it.

\subsection{Related works}\label{Related_Works}
Machine learning methods have been used for ECG analysis in a variety of applications. There has been a wealth of work in the classification of various Cardiovascular Diseases (CVDs) from ECG data~\cite{vemishetty2019phase,vemishetty2016classification,rocha2008phase,roberts2001identification, zhang2020ecg, pourbabaee2018deep,kiranyaz2015real,fan2018multiscaled, lih2020comprehensive}. Other applications of machine learning in ECG analysis include detecting seizures and heart attacks~\cite{lee2014classification,liu2017real}, predicting patients' blood pressure~\cite{miao2020continuous}, detecting a patients facial expressions~\cite{dawid2019psr} and analysis of ECG of the brain has been used for creating brain computer interfaces (BCI) capable of detecting which body part the subject was completing a task with~\cite{djemal2016three,chen2014phase}.

A popular technique for preprocessing ECG data is to create its Phase Space Reconstruction (PSR) matrix. Typically, features are extracted from the PSR matrix of ECG data which can then be used as inputs for a classification model. Box counting as well as column and row statistics are features often extracted from the PSR matrix of ECG data. These methods have been used in the prediction of CVD~\cite{vemishetty2019phase,vemishetty2016classification,rocha2008phase,roberts2001identification}, creating BCIs~\cite{chen2014phase,djemal2016three}, and detecting facial expressions~\cite{dawid2019psr}. These approaches all centre around manually selecting features to extract from the PSR matrix. Our proposed method diverges from this by using the whole PSR matrix as the input to a model, which is itself, capable of extracting features. Convolutional neural networks (CNN) are an example of this. During training, convolutional layers learn to extract features of the input image which are most impactful in accurately determining the models output. As such, these models can replace the need for time consuming feature extraction and can arrive at much more descriptive features. CNNs have been used in ECG analysis for classifying heart attacks~\cite{liu2017real}, atrial fibrillation~\cite{fan2018multiscaled,pourbabaee2018deep}, and other arrhythmias~\cite{kiranyaz2015real,zhang2020ecg,sangaiah2020intelligent, lih2020comprehensive} as well as for predicting blood pressure~\cite{miao2020continuous}. CNN are typically used for the classification of 2D images and, as such, use 2D filters for feature extraction. All of these methods use the filtered ECG as the model input, where each lead corresponds to a single 1D signal. By creating the PSR matrix for each lead, we create a 2D input from each lead.

\subsection{Contributions and outline of the paper}

In this paper, we propose an accurate, reliable, and reproducible method that utilises the concept of prolonged screening for S-ICD eligibility to better scrutinise the selection criteria in an attempt to find a cohort of patients with low probability of TWOS and inappropriate shocks. Central to this prolonged screening is a deep learning based method for predicting the T:R ratio---the main determinant of S-ICD eligibility---of a 10-second segment of a single lead ECG. Solving this regression problem is at the core of this paper. The aim of our proposed screening method is to use the model developed in this paper to analyse the T:R ratio of each single lead 10-second ECG segment within a three lead 24-hour ECG recording, to determine if any of the three leads have a suitably low risk of TWOS. For a given lead, should a patient have a T:R ratio above 1:3 for a continuous period of at least 20 seconds (the duration of TWOS at which the S-ICD would deliver an inappropriate shock), that lead would fail the screening. If all three leads fail the screening the patient would be deemed not eligible for an S-ICD. Temporal variations in the T:R ratio make the current 10-second screening process unreliable. Our proposed method allows for a much more robust screening, as it would allow for the analysis of variations in the T:R ratio over a 24-hour period. Subsection ~\ref{screening_example} gives an example of how the model for predicting T:R ratio from ECG segments, developed in this paper, could be used within a screening tool to access patients' eligibility for S-ICD implantation.

In contrast with most of the literature, in which, with only a few exceptions (see, e.g.,~\cite{babu2016deep}), CNNs are used for image classification rather than for regression, we propose using 2D PSR images of the filtered ECG signals as input to CNNs with 2D filters.
In particular, rather than extracting features from PSR images which are then used as model inputs, as typically done in previous works, we use the entire PSR image as input to our models.
To the best of our knowledge, this has not yet been done in the field of ECG analysis---while CNNs have been used to analyse ECGs, this was only done using 1D-ECG signals as inputs to CNNs with 1D filters.

In Section~\ref{Methodology}, we outline the preprocessing techniques used to create PSR images of the filtered 10-second ECG segments, detail our selection of deep learning model architectures and describe the process we use to train and evaluate these models. Section~\ref{Experiments}, illustrates the results of an extensive set of experiments, demonstrating the capabilities of our models to predict T:R ratios from PSR images as well as providing examples of how such models could be integrated into a clinical tool. In Section~\ref{summary}, we summarise the paper and outline directions for future work.

\section{Methodology}\label{Methodology}

We propose a new screening process, wherein a Holter\textregistered---a portable ECG recording device---is used to record 24 hours of ECG data from three leads corresponding to the three vectors utilised by the S-ICD. This data is then split into 10-second segments and an artificial neural-network based model is used to predict the T:R ratio for each 10-second segment. The cardiologist would then be able to review the behavior of the patient's T:R ratio over the 24-hour period and evaluate their eligibility. We rely on a number of filtering techniques for removing noise from the ECG signals. We then use PSR, a popular technique in waveform analysis, to convert the ECG signal into an image of it's PSR matrix with which we then train a Convolutional Neural Network (CNN) to predict the T:R ratio from these images.

The most straightforward approach to measuring the T:R ratio is to locate the R and T waves and measure their amplitudes. However, by explicitly detecting and measuring the peaks of the R and T waves, we run the risk of TWOS when the characteristics of the T wave becomes similar to those of the R wave. To avoid this, our model aims to predict the T:R ratio without ever explicitly locating or measuring the R or T waves.

\begin{figure*}[t]
  \centering
  \includegraphics[width=\textwidth]{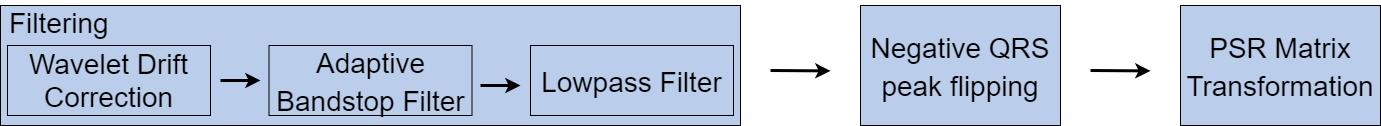}
  \caption{Flowchart of the methods in our data preprocessing step.}\label{Preprocessing_flowchart}
\end{figure*}

\subsection{Preprocessing}\label{Preprocessing}
Preprocessing involves filtering the ECG data, performing transformations to emulate the methods used within S-ICDs and creating images by plotting the PSR of the data. Figure~\ref{Preprocessing_flowchart} gives an overview of the preprocessing techniques used to prepare our data for the training of the regression models.

\subsubsection{Data Structure}\label{DataStructure}

For this paper, we consider data in the form of 10-second segments of single lead ECG recordings from Holter leads corresponding to the vectors used by the S-ICD.
These 10-second ECG segments are annotated with the positions of the peaks of the T and R waves occurring in this period.

From these annotations, we are able to calculate the dependant variable for our regression problem: the average T:R ratio.
As mentioned previously, the T:R ratio of a single PQRST complex is simply the ratio between the amplitudes of the T and R waves.
For the purposes of this paper, we will consider this ratio in the form of a fraction.
The average T:R ratio for a 10-second ECG segment $\{x(1), \ldots, x(10\cdot f)\}$, where $f$ is the sampling frequency of the signal, with T-peak annotations at indexes $\{T_1, \ldots, T_n\}$ and R-peak annotations at indexes $\{R_1, \ldots, R_n\}$ is given by

\begin{equation}\label{AverageTRRatio}
\frac{\sum_{i=1}^{n}x(T_i)}{\sum_{i=1}^{n}x(R_i)}.
\end{equation}

When the T wave has negative amplitude, this fraction is negative. From a clinical perspective, we are only interested in the magnitude of the T:R ratio.
However, from a signal processing perspective, there is a great difference between a PQRST complex with a negative T wave and one with a positive T wave.
For this reason, we will preserve the sign of the T:R ratio, as the loss of information resulting from considering only the magnitude of the ratio would lead to a reduction in the accuracy of our models.
Having built a model capable of predicting T:R ratios from 10-second ECG segments, we take the magnitude of this model's outputs for use in a clinical tool.
Our choice to consider the T:R ratio rather than the R:T ratio, which is more common in the literature, is well motivated.
As the T wave amplitude approaches $0$, very small changes in the T wave amplitude can result in extreme changes in the R:T ratio.
This massive variation in R:T ratio for very similar ECG signals makes it inappropriate for use as a label in our regression problem.
Typically, the R wave is of greater amplitude than the T wave. Because of this, the T:R ratios of a set of ECG segments are well distributed between $0$ and $1$.
For this reason, we use the T:R ratio as our dependent variable in our regression problem.
If the situation requires it, the R:T ratio can of course be derived from our model by simply taking the inverse of the model's output.

\subsubsection{Filtering}\label{Filtering}

Figure \ref{Preprocessing_flowchart} gives an overview of the filtering methods we will use to remove noise from our ECG signal \cite{lugovaya2005biometric}.

Firstly, baseline drift correction is implemented using one-dimensional Discrete Wavelet Transformation (DWT). The ECG signal is decomposed at 9 levels, using the Daubechies 8 (db8) wavelet, then reconstructed using only level 9 coefficients. This reconstructed signal is the low frequency component for the ECG signal which is assumed to be the drifting baseline. Subtracting this from the original signal leaves us with an ECG signal with a stable baseline of value 0.

Adaptive bandstop filtering is used to suppress power-line noise with a frequency of 50Hz, while a lowpass filter is used to remove the remaining high-frequency noise. Having applied these filters, the locations of the R and T peak markers may no longer be correct. To account for this, a small region around the R peak is searched for a maximum and this maximum is taken as the new R peak. Similarly, a small region around the T peak is searched and the maximum or minimum value in this region is taken as the new T peak if the T peak is positive or negative, respectively.

\begin{figure}[h]
  \centering
  \includegraphics[width=\columnwidth]{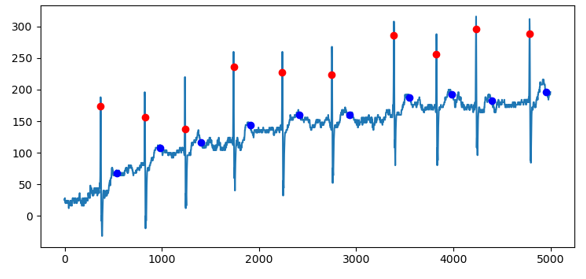}
  \caption{Example of a 10-second ECG segment pre-filtering. R and T peaks shown in red and blue, respectively.}\label{FilteringExampleUnfiltered}
\end{figure}

\begin{figure}[h]
  \centering
  \includegraphics[width=\columnwidth]{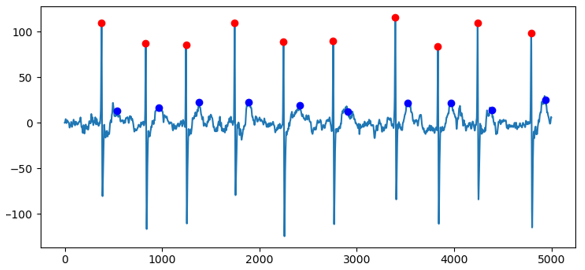}
  \caption{The 10-second ECG segment shown in Figure \ref{FilteringExampleUnfiltered} post-filtering. R and T peak
  s shown in red and blue, respectively.}\label{FilteringExampleFiltered}
\end{figure}

Figure \ref{FilteringExampleUnfiltered} gives an example of an unfiltered 10-second ECG segment, with the filtered signal shown in Figure \ref{FilteringExampleFiltered}.


\subsubsection{Negative QRS peak flipping}\label{Filtering}
\begin{figure}[ht]
  \begin{tabular}{cc}
    \includegraphics[scale=0.15]{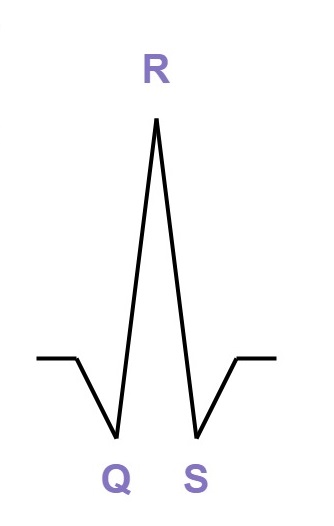} &   \includegraphics[scale=0.15]{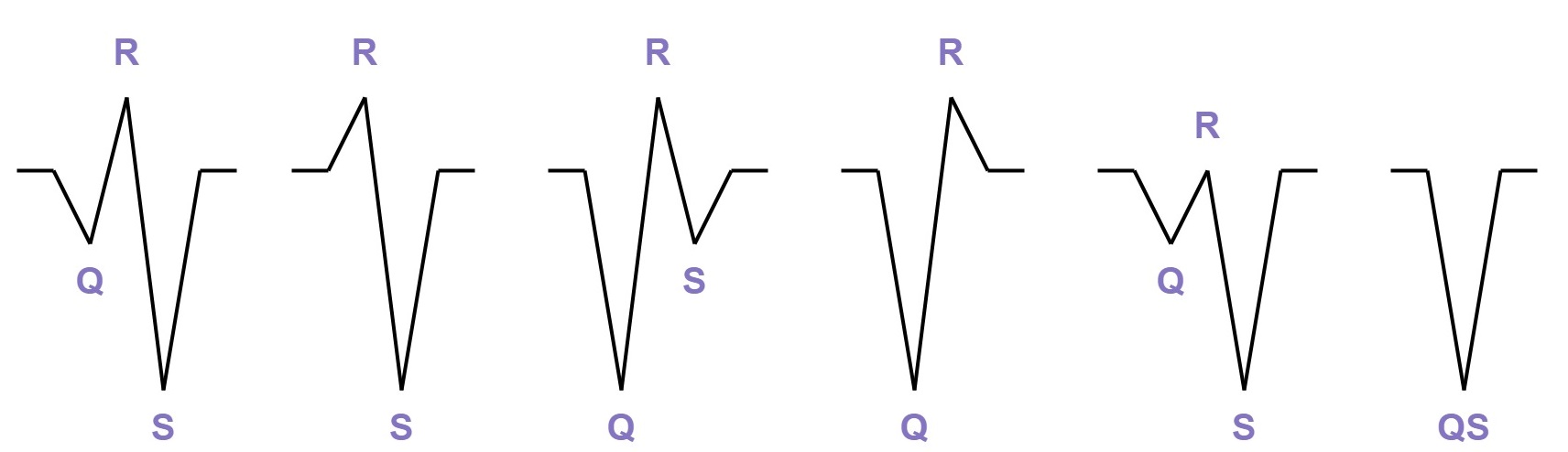}
    \\
    (a) & (b)
  \end{tabular}
  \caption{(a) An example of a normal QRS complex. (b) Examples of QRS complexes in which the R wave is not the wave of greatest amplitude.}
  \label{negative_r_waves}
\end{figure}

While R waves are strictly positive, a PQRST complex with a small R wave could be prone to the T wave being labeled as an R wave, leading to double counting. For this reason, when implementing R peak detection, it may be reasonable to look for both positive and negative peaks and simply flip the ECG when the peak detected is negative. This approach is followed by the algorithm used in S-ICDs. Figure \ref{negative_r_waves} shows some possible QRS complexes where the amplitude of the Q or the S wave is greater than that of the R wave.

In cases where the peak of greatest magnitude of a QRS complex is negative and it is labelled as the R peak, we are, in actuality, labeling either the Q or the S peak. In this case, the ECG signal of the entire PQRST complex is flipped (multiplied by $-1$) and the R peak is assigned as the previously negative peak.
To implement this on our data, we search within a region of each of the R peak markers for a negative peak of greater magnitude. If one is found, it is assigned as the new R peak and the signal is flipped.
\begin{figure}[H]
  \centering
  \includegraphics[width=\columnwidth]{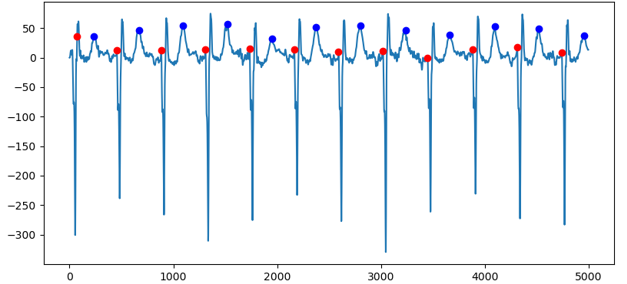}
  \caption{Example of a filtered 10-second ECG segment. R and T peaks shown in red and blue, respectively.}\label{NegativeFlipExampleUnflipped}
\end{figure}
\begin{figure}[H]
  \centering
  \includegraphics[width=\columnwidth]{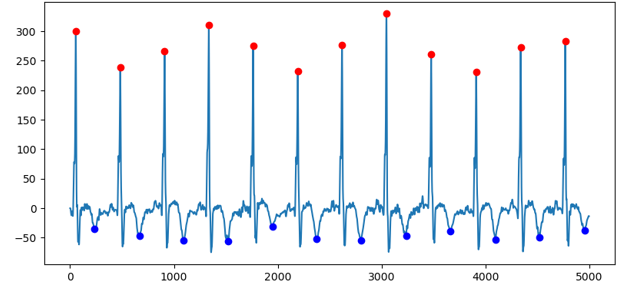}
  \caption{The 10-second ECG segment shown in Figure \ref{NegativeFlipExampleUnflipped} post-flipping. R and T peaks shown in red and blue, respectively.}\label{NegativeFlipExampleFlipped}
\end{figure}
Figure \ref{NegativeFlipExampleUnflipped} gives an example of a filtered 10-second ECG segment while Figure \ref{NegativeFlipExampleFlipped} shows the same segment after the negative QRS peaks have been detected, the signal has been flipped and the peaks have been reassigned.

\subsubsection{Phase space reconstruction}\label{PSR}
Phase space reconstruction is a technique for representing non-linear characteristics of a time series set of data using delay maps. For a given time series $x(1), x(2), \ldots, x(n)$, the time lagged phase space vectors are given by
\begin{equation}
\begin{aligned}
&X(i) = \left\{x(i), x(i+\tau), \ldots , x(i+(d-1)\tau)\right\} \\
&\quad\quad \text{for }i \in \{1 , \ldots, n-(d-1)\tau\},
\end{aligned}
\end{equation}
where $\tau$ is the time delay between points in the series and $d$ is the number of dimensions of the phase space which we are mapping this data to. While using high dimensional PSR has given good results in the field of BCI~\cite{chen2014phase,djemal2016three}, the majority of the work in ECG analysis uses two-dimensional PSR, corresponding to $d=2$~\cite{rocha2008phase,krishnan2007phase,vemishetty2019phase,roberts2001identification}. For this reason we transform our time series data into a normalized matrix of two-dimensional phase space vectors

\begin{equation}
\begin{aligned}
&B = \begin{bmatrix}
x(1)/q & x(1+\tau)/q\\
x(2)/q & x(2+\tau)/q\\
\vdots & \vdots\\
x(n-\tau)/q & x(n)/q
\end{bmatrix},\\
&\text{where }\quad q = \max\left\{|x(i)|:  i = 1, \ldots, n\right\}.\\
\end{aligned}
\end{equation}

In two dimensions, the phase space plot corresponding to this matrix ranges from $-1$ to $+1$ on each axis. This area can now be divided into $N^2$ small square areas, $g(i,j)$, of size $R\times R$ for $i,j=1,\ldots,N$, where $R$ is given by $R = 2/N$ and $N$ is an integer number. The phase space matrix $C$ (of dimension $N\times N$) is constructed with each of its elements $c(i,j)$ equal to the number of phase space vectors in $B$ which fall within the square area $g(i,j)$. $C$ is then normalised to give $P$, wherein each element $p(i,j)$ gives the probability of a phase point falling within $g(i,j)$. Formally,

\begin{figure*}[b]\label{N_Deep_ANN}
  \centering
  \includegraphics[width=\textwidth]{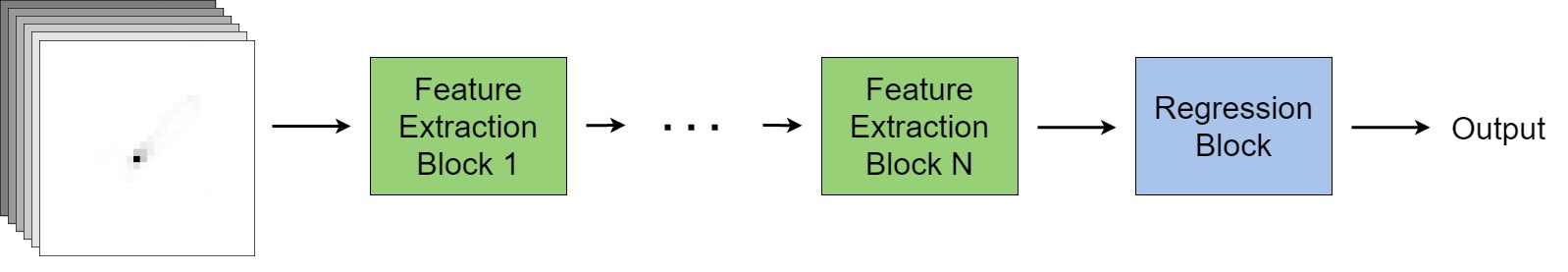}
  \caption{Diagram of a model comprised of N feature-extraction blocks}\label{N_Deep_ANN}
\end{figure*}

\begin{equation}\label{P}
\begin{aligned}
&P = \frac{1}{M}C, &&M = \sum_{i=1}^{N}\sum_{j=1}^{N}c(i,j).
\end{aligned}
\end{equation}

A typical approach is to extract features from these PSR by either box-counting~\cite{vemishetty2019phase}, calculating the spatial filling index~\cite{krishnan2007phase} or calculating statistics of the distributions of values within each column of $C$~\cite{rocha2008phase}. These features are then used as model inputs for classifying various different categories of ECG. In this paper, we use tools typically used for computer vision to allow us to use the $N\times N$ pixel images derived from $P$ as the input for our model.

\begin{figure}[H]
  \begin{tabular}{cc}
    \includegraphics[scale=0.36]{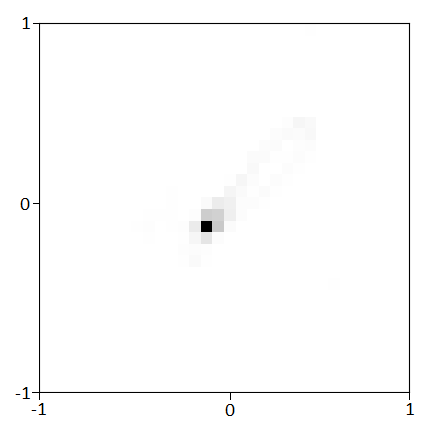} &   \includegraphics[scale=0.36]{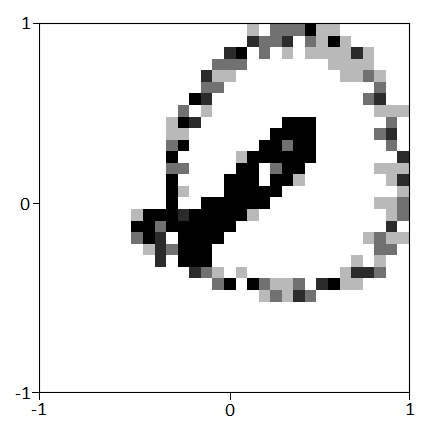}
    \\
    (a) & (b)
  \end{tabular}
  \caption{(a) Image of the PSR matrix formed from a 10-second segment ECG signal with $N=32$. (b) Darkened image of the same PSR matrix.}
  \label{PSR_Images}
\end{figure}

Figure \ref{PSR_Images} gives an example of a PSR image with $N=32$ as well as a darkened version of the same image for readability. As one can see in the darkened image, there are some connected bands of higher probability. With values of $N$ greater than 32, these bands become disconnected as no phase space vectors land within that portion of the grid.

\subsection{MLP and CNN for regression}\label{CNN_for_Regression}
In this subsection, we discuss the architecture of the models used to predict T:R ratios from 32x32 pixel PSR images. The general structure of each model is laid out in Figure \ref{N_Deep_ANN}. Each model is made up of N feature-extraction blocks, followed by a regression block.

The regression block is used by all models to derive the T:R ratio from the extracted features. The outputs from the preceding feature-extraction blocks are flattened to a 1D vector and fed into a series of fully connected (dense) layers of neurons to arrive at the final regression output: the T:R ratio. Table~\ref{Regression Block} gives an overview of the layers comprising this block. We use batch normalisation for regularisation as it has been shown to be superior to dropout for use in CNNs~\cite{garbin2020dropout}. We perform batch normalisation before applying the activation function as proposed in the original paper on batch normalisation~\cite{ioffe2015batch}.

\begin{table}[h!]
\caption{Regression Block}
\small
\centering
\setlength{\tabcolsep}{3pt}
\begin{tabular}{lll}
\hline
Layer & Type & Output size\\
\hline
1  &Dense	&256\\
2  &BatchNorm	&256\\
3  &Activation(Relu)	&256\\
4  &Dense	&64\\
5  &BatchNorm	&64\\
6  &Activation(Relu)	&64\\
7  &Dense	&1\\
\hline
\end{tabular}
\label{Regression Block}
\end{table}

The first and most basic of our feature-extraction blocks is the MLP feature-extraction block, comprised of a single layer of fully connected neurons followed by a batch normalisation and activation layer. The input and output of these blocks are 1D. As such, when using these blocks, we flatten the images before the first feature-extraction block rather than before the regression block.

\begin{table}[h!]
\caption{MLP Feature Extraction Block n}
\small
\centering
\setlength{\tabcolsep}{3pt}
\begin{tabular}{lll}
\hline
Layer & Type & Output size\\
\hline
1  &Dense	&1024\\
2  &BatchNorm	&1024\\
3  &Activation(Relu)	&1024\\
\hline
\end{tabular}
\label{MLP Feature Extraction Block n}
\end{table}

The basic CNN feature-extraction blocks utilise convolutional layers, which exploit the 2D structure of the PSR images, as opposed to fully connected layers. These layers are followed by the batch normalisation and activation layers mentioned previously and finally a maximum pooling layer to reduce the size of the output images. As shown in Table \ref{Basic CNN Feature Extraction Block n}, the output of each layer and the kernel size of the convolutional layer depends on the number of feature-extraction blocks.

\begin{table}[h!]
\caption{Basic CNN Feature Extraction Block n}
\small
\centering
\setlength{\tabcolsep}{3pt}
\begin{tabular}{lllll}
\hline
Layer & Type & Output size & Kernel size & Stride\\
\hline
1  &Convolutional	&$2^{n+4}$x$2^{6-n}$x$2^{6-n}$ &$(7-n)x(7-n)$ &1\\
2  &BatchNorm	&$2^{n+4}$x$2^{6-n}$x$2^{6-n}$\\
3  &Activation(Relu)	&$2^{n+4}$x$2^{6-n}$x$2^{6-n}$\\
4  &MaxPooling &$2^{n+4}$x$2^{5-n}$x$2^{5-n}$ &2x2 &2\\
\hline
\end{tabular}
\label{Basic CNN Feature Extraction Block n}
\end{table}

The complex CNN feature-extraction block is based on the basic CNN feature-extraction block. The convolutional layer is replaced by a pair of convolutional layers with smaller kernels. The maximum pooling layer is replaced by an additional convolutional layer with stride = 2, which reduces the size of the output whilst continuing to extract features. Finally, addition skip connections, as utilised in ResNet \cite{he2016deep}, are added over the first two convolutional layers in order to speed up training. Table \ref{Complex CNN Feature Extraction Block n} gives an overview of the layers of this block.

\begin{table}[h!]
\caption{Complex CNN Feature Extraction Block n}
\small
\centering
\setlength{\tabcolsep}{3pt}
\begin{tabular}{lllll}
\hline
Layer & Type & Output size & Kernel size & Stride\\
\hline
1  &Convolutional	&$2^{n+2}$x$2^{6-n}$x$2^{6-n}$ &$(6-n)x(6-n)$ &1\\
2  &BatchNorm	&$2^{n+2}$x$2^{6-n}$x$2^{6-n}$\\
3  &Activation(Relu)	&$2^{n+2}$x$2^{6-n}$x$2^{6-n}$\\
4  &Convolutional	&$2^{n+2}$x$2^{6-n}$x$2^{6-n}$ &$(6-n)x(6-n)$ &1\\
5  &BatchNorm	&$2^{n+2}$x$2^{6-n}$x$2^{6-n}$\\
6  &Activation(Relu)	&$2^{n+2}$x$2^{6-n}$x$2^{6-n}$\\
7  &Skip(Input)	&$2^{n+2}$x$2^{6-n}$x$2^{6-n}$\\
8  &Convolutional &$2^{n+3}$x$2^{5-n}$x$2^{5-n}$ &$(7-n)x(7-n)$ &2\\
9  &BatchNorm &$2^{n+3}$x$2^{5-n}$x$2^{5-n}$\\
10 &Activation(Relu) &$2^{n+3}$x$2^{5-n}$x$2^{5-n}$\\
\hline
\end{tabular}
\label{Complex CNN Feature Extraction Block n}
\end{table}

The Deep CNN feature-extraction block is very similar to the complex CNN feature-extraction block. Before the convolutional layer with stride = 2 which is used for pooling, we include an additional pair of convolutional layers with smaller kernels as well as a second skip connection. Table \ref{Deep CNN Feature Extraction Block n} gives an overview of the layers of this block.

\begin{table}[h!]
\caption{Deep CNN Feature Extraction Block n}
\small
\centering
\setlength{\tabcolsep}{3pt}
\begin{tabular}{lllll}
\hline
Layer & Type & Output size & Kernel size & Stride\\
\hline
1  &Convolutional	&$2^{n+2}$x$2^{6-n}$x$2^{6-n}$ &$(6-n)x(6-n)$ &1\\
2  &BatchNorm	&$2^{n+2}$x$2^{6-n}$x$2^{6-n}$\\
3  &Activation(Relu)	&$2^{n+2}$x$2^{6-n}$x$2^{6-n}$\\
4  &Convolutional	&$2^{n+2}$x$2^{6-n}$x$2^{6-n}$ &$(6-n)x(6-n)$ &1\\
5  &BatchNorm	&$2^{n+2}$x$2^{6-n}$x$2^{6-n}$\\
6  &Activation(Relu)	&$2^{n+2}$x$2^{6-n}$x$2^{6-n}$\\
7  &Skip 1(Input)	&$2^{n+2}$x$2^{6-n}$x$2^{6-n}$\\
8  &Convolutional	&$2^{n+2}$x$2^{6-n}$x$2^{6-n}$ &$(6-n)x(6-n)$ &1\\
9  &BatchNorm	&$2^{n+2}$x$2^{6-n}$x$2^{6-n}$\\
10  &Activation(Relu)	&$2^{n+2}$x$2^{6-n}$x$2^{6-n}$\\
11  &Convolutional	&$2^{n+2}$x$2^{6-n}$x$2^{6-n}$ &$(6-n)x(6-n)$ &1\\
12  &BatchNorm	&$2^{n+2}$x$2^{6-n}$x$2^{6-n}$\\
13  &Activation(Relu)	&$2^{n+2}$x$2^{6-n}$x$2^{6-n}$\\
14  &Skip 2(Skip 1)	&$2^{n+2}$x$2^{6-n}$x$2^{6-n}$\\
15  &Convolutional &$2^{n+3}$x$2^{5-n}$x$2^{5-n}$ &$(7-n)x(7-n)$ &2\\
16  &BatchNorm &$2^{n+3}$x$2^{5-n}$x$2^{5-n}$\\
17 &Activation(Relu) &$2^{n+3}$x$2^{5-n}$x$2^{5-n}$\\
\hline
\end{tabular}
\label{Deep CNN Feature Extraction Block n}
\end{table}
When referring to the model, we give the type of feature-extraction block used followed by the number of feature-extraction blocks. For example, a model comprised of 5 MLP feature-extraction blocks followed by the regression block would be referred to as MLP5, while a model comprised of 3 Complex CNN feature-extraction blocks followed by the regression block would be referred to as Complex CNN3.

\subsection{Model enhancement}\label{Model_Augmentation}

\begin{figure}[h!]
  \centering
  \includegraphics[width=.15\textwidth]{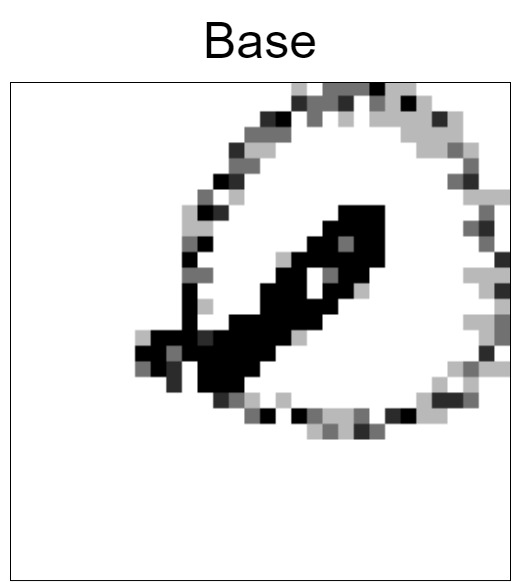}
  \includegraphics[width=.15\textwidth]{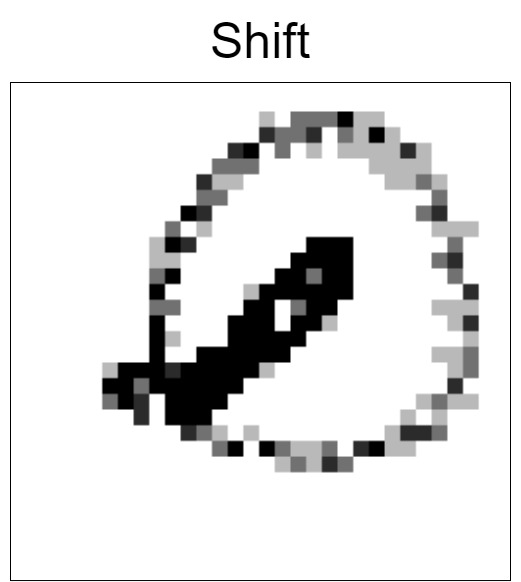}
  \includegraphics[width=.15\textwidth]{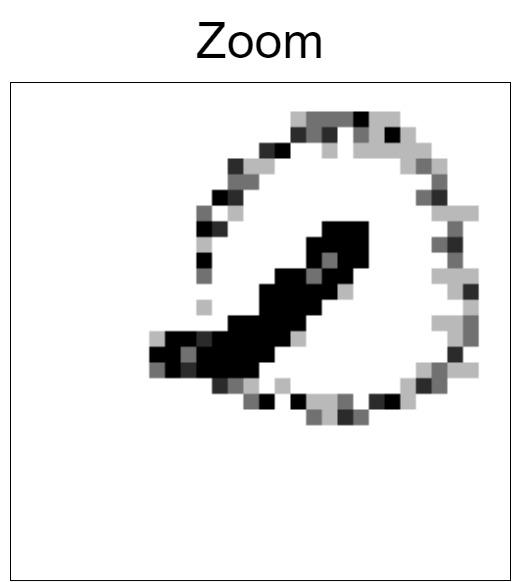}
  \\[\smallskipamount]
  \includegraphics[width=.15\textwidth]{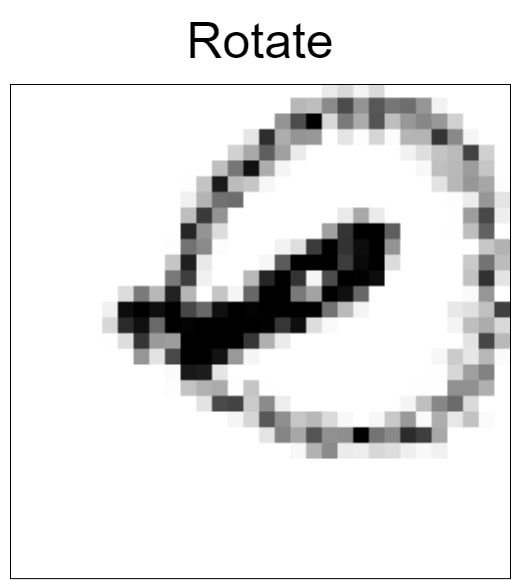}
  \includegraphics[width=.15\textwidth]{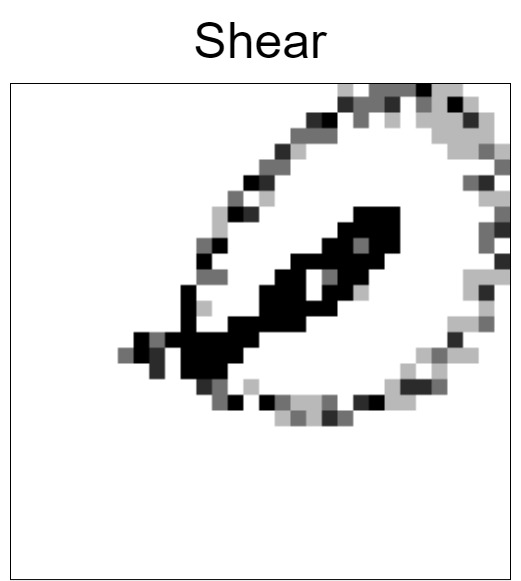}
  \caption{Examples of how shifting down and to the left by 12.5\%, zooming by -12.5\%, shearing with an angle of 15$^{\circ}$ and rotating by 15$^{\circ}$ effect a PSR image.}\label{ImageAug}
\end{figure}

To improve the performance of our models, described in the previous subsection, we consider following two techniques.
Firstly, image augmentation, which can strengthen the training of models by randomly introducing small distortions to the training data, thus resulting in models which are more robust to distortions which may occur naturally in our unaltered data.
There are many image augmentation strategies, however only a small number of them are appropriate for the sort of image we are handling. We test 4 image augmentation strategies: shifting, zooming, rotating and shearing. Figure \ref{ImageAug} gives an example of how each of these augmentations could affect a PSR image.
When implementing image augmentation, a model may be trained on data which, at each mini-batch, is either shifted randomly within a range of $[1-\phi/100, 1+\phi/100]$, zoomed randomly within a range of $[1-\phi/100, 1+\phi/100]$, rotated randomly by an angle within a range of $[-\phi^{\circ}, +\phi^{\circ}]$ or sheared randomly with a shear angle within a range of $[-\phi^{\circ}, +\phi^{\circ}]$. We denote the augmentation as either tiny $(\phi = 2.5$), small ($\phi = 5$) or moderate ($\phi = 10$).
For instance, MLP5 Small Rotation would refer to a model consisting of 5 MLP feature- extraction blocks followed by a regression block, which has been trained on images which have been, at each mini-batch, rotated randomly by an angle within a range of $[-5^{\circ},5^{\circ}]$.
%

Secondly, we consider ensemble models.
In order to avoid overfitting, 20\% of our training data is randomly reserved for validation.
Models are trained until their accuracy on the validation set is no longer increasing. This means that 20\% of our original training data is not being used for training.
To avoid this, we can iteratively reserve a different 20\% of our training data for validation, training 5 models with the same architecture but which have all used a different set of data for validation.
In doing so, while some data may be used for validation by one model, it will be used for training by the remaining 4.
By taking the average of these models' predictions, we aim to obtain a higher prediction accuracy than would be archived by any single model.
When referring to an ensemble model comprised on 5 sub-models, we simply append the word ensemble onto the name of the sub-model.

\begin{figure*}[!b]
  \centering
  \includegraphics[width=16cm]{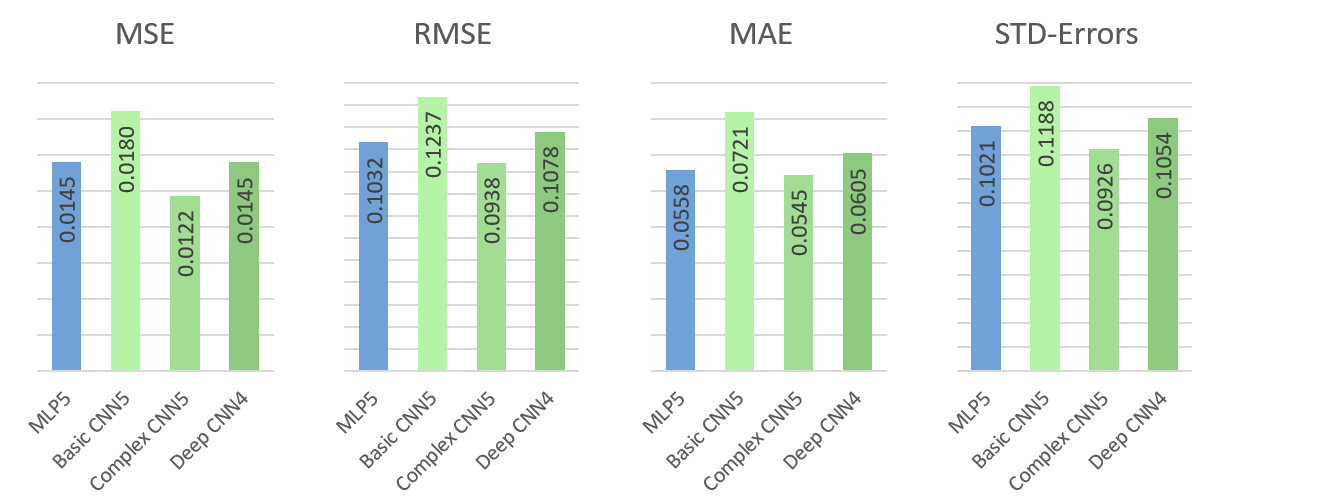}
  \caption{Performances of the best models using each type of feature extraction block}\label{Experiment1Results}
\end{figure*}

\subsection{Model evaluation}\label{Model_Evaluation}
To evaluate these models, we use 10-fold cross validation. At each iteration we create the following splits in our data: 10\% of the data is used as the testing set; for ultimately evaluating the models we train, 20\% of the remaining data forms the validation set; to help avoid overfilling during training, the remaining data is used to form the training set. We assess the accuracy of each model using mean squared error (MSE), root mean squared error (RMSE), mean absolute error (MAE) and STD of errors respectively defined as
\begin{equation}\label{Accuracy Metrics}
\begin{aligned}
  &\text{MSE} = \frac{\sum_{i = 1}^{n}(y_i - \hat{y}_i)^2}{n},\\
  &\text{RMSE} = \sqrt{\frac{\sum_{i = 1}^{n}(y_i - \hat{y}_i)^2}{n}}, \\
  &\text{MAE} = \frac{\sum_{i = 1}^{n}|y_i - \hat{y}_i|}{n}, \\
  &\text{STD of errors} = \sqrt{\frac{\sum_{i = 1}^{n}\left(y_i - \hat{y}_i - \frac{\sum_{j = 1}^{n}y_j - \hat{y}_j}{n}\right)^2}{n-1}},
\end{aligned}
\end{equation}
where $y$ are the true T:R ratios, $\hat{y}$ are the predicted T:R ratios and $n$ is the number of PSR images.

\section{Experiments}\label{Experiments}


\subsection{SGH-ECG and ECG-ID data sets}\label{Data_Sets}
The primary data set we have used in this study was collected at Southampton General Hospital as part of a study on S-ICD eligibility. We refer to this data set as the Southampton General Hospital ECG (SGH-ECG) data set. The SGH-ECG data set consists of 390 10-second ECG segments, sampled at 500 Hz, annotated with R and T peaks. These signals were obtained at random intervals from the 24-hours ECG recordings of 18 different participants. \footnote{This study was performed with favourable opinion from the REC (17/SC/0623) and with R\&D (RHMCAR0528) approval. This study was conducted in accordance with the Research Governance Framework for Health and Social Care (2005), Good Clinical Practice and their relevant updates.} Eight of the 18 participants have underlying congestive heart failure, 3 participants have underlying congenital heart disease, 3 have ``hypertrophic cardiomyopathy'' and 4 participants have a structurally normal heart. The participants ages range from 20 to 80 years old with a mean of 53.16 years. There is an even split of male and female participants. Using the preprocessing methods laid out in Subsection \ref{Preprocessing}, 32x32-pixel PSR images and their corresponding T:R ratios are derived from these 10 ECG segments. Our aim is to build a model capable of accurately predicting T:R ratios from these PSR images.

In order to increase the amount of training data, at each round of cross validation, after testing and validation sets have been reserved, we bolster our training set by combining it with the 32x32 pixel PSR images and T:R ratios derived from the ECG Identification (ECG-ID) Database. The ECG-ID data set, collected by Lugovaya~\cite{lugovaya2005biometric}, consists of 310 20-second ECG segments sampled at 500Hz, with R and T peak annotations for the first 10 heartbeats. These signals are obtained from 90 participants, 44 men and 46 women, with ages ranging from 13 to 75.

\subsection{Experiment 1: architecture selection}\label{architecture_selection}
Our first experiment is to assess which model architectures, as detailed in Subsection \ref{CNN_for_Regression}, most accurately predict T:R ratios from 32x32 pixel PSR images. Figure \ref{Experiment1Results} shows the average accuracy across each of the 10 rounds of cross validation of some of the best performing models using the accuracy metrics laid out in Subsection \ref{Model_Evaluation} and the model naming convention laid out in Subsection \ref{CNN_for_Regression}.
\begin{figure*}[!b]
  \centering
  \includegraphics[width=16cm]{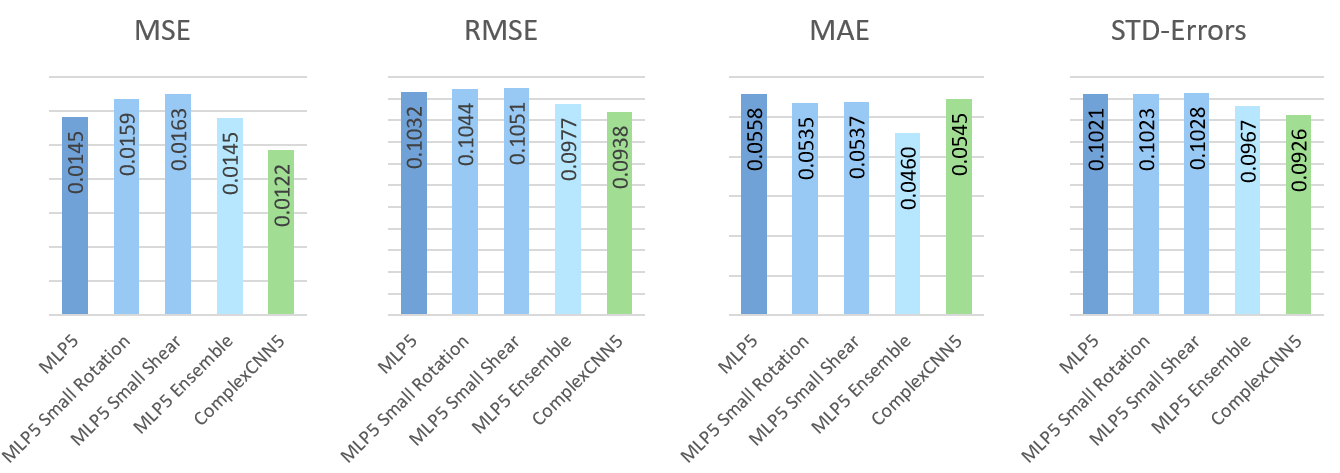}
  \caption{\footnotesize Comparison between MLP5 model variants and the Complex CNN5 model}\label{Experiment2Results}
\end{figure*}




As can be seen from the results in Figure \ref{Experiment1Results}, the MLP5 model is capable of accurately predicting T:R ratios with a mean absolute prediction error of only 0.0558. (Recall, T:R ratios range between 0 and 1 and the threshold for failing a screening is 0.33)
In switching to a Basic CNN structure, we see a considerable drop in accuracy. Adding more convolutional layers by using Complex CNN feature-extraction blocks, allows us to recoup this loss.
The Complex CNN5 model outperforms the MLP5 model for all tested metrics.
However, continuing to add convolutional layers, by using Deep CNN feature-extraction blocks results in a drop in accuracy for all tested metrics.
This is likely due to the fact that our data set is not sufficiently large to enable the training of such deep neural-networks.

Figure \ref{Experiment1Results} shows the average cross validation accuracies of only the best performing model for each type of feature-extraction block.
There are several models, with architectures utilising the MLP and Complex CNN feature-extraction blocks, with a MAE under 0.06 indicating that on average these models are able to predict T:R ratios within 0.06 of their true value.

\subsection{Experiment 2: model enhancement}\label{model enhancement}
In this Experiment, we test the effect of image augmentation and creating ensemble models on our two best performing models from the previous experiment: MLP5 and Complex CNN5.
We test 4 different image augmentation schemes (shifting, zooming, rotating and shearing) at three different magnitudes (tiny, small and moderate).
We also evaluate ensemble models, created by averaging the prediction of 5 sub-models which each use a different portion of the training set for validation.
The details of both of these methods are given in Subsection \ref{Model_Augmentation}

Neither method for model enhancement had a significant positive impact on the accuracies of the Complex CNN5 model.
For this reason, Figure \ref{Experiment2Results} compares the average accuracy across the 10 rounds of cross validation of the enhanced MLP5 models which were able to outperform the base MLP5 model with that of the MLP5 and the Complex CNN5 models.
%


The only two image augmentation schemes which improve the performance of the MLP5 model are small rotations and small shears, which both only result in small increases in some accuracy metrics, along with small decreases in others.
The MLP5 Ensemble model gives modest improvements in performance, outperforming all other variants of the MLP5 model on each metric.
However, even the MLP5 Ensemble model is outperformed by the Complex CNN5 model on every metric aside from MAE, where the MLP5 Ensemble model is able to achieve an error of only 0.460.

As MSE and RMSE are generally considered better measures of the regression capabilities of a model than MAE, we determine that the Complex CNN5 model is the best performing regression model as it significantly outperforms all tested models on these metrics.
An additional drawback to the MLP5 Ensemble model is that, while the MLP5 model is less computationally expensive to train than the Complex CNN5 model, due to its more simplistic architecture, training the 5 MLP5 models required for creating the MLP5 Ensemble model is more expensive than training a single Complex CNN5 model.

Although, individually, rotating and shearing results in improved performance for the MLP5 model, we find that using both of these image augmentation schemes in concert does not result in improved performance. Likewise, we do not observe any impartment in performance  when using image augmentation whilst creating ensemble models.

\subsection{Screening tool}\label{screening_example}
The previous two experiments have shown that the Complex CNN5 model outperforms all other tested models and is capable of predicting T:R ratios within 0.0545 of their true value with an average MSE of only 0.122.
In this section, we illustrate how this model could be used for the purposes of screening patients for S-ICD implantation eligibility.

\begin{figure*}[!hb]
  \centering
  \includegraphics[width=16cm]{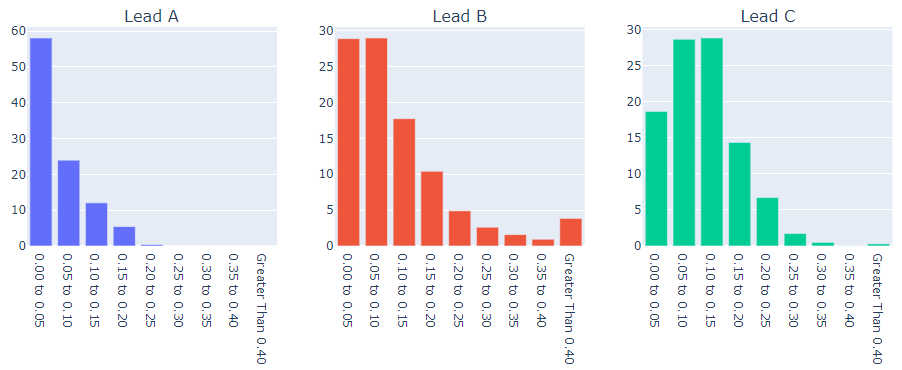}
  \caption{Histogram of the predicted TR ratio over the 24-hour screening preriod for each lead}\label{ToolHistograms}
\end{figure*}
After recording 24-hour ECG signals on multiple leads, each continuous 24-hour signal would be broken into 8640 non-overlapping 10-second segments which would then be processed using the methodology detailed in Subsection \ref{Preprocessing}, resulting in 8640 PSR images for each lead, which are ordered chronologically.
We would then input these images to our model and produce predictions for the T:R ratio of each 10-second segment represented by a single PSR image.
As mentioned previously, from a clinical perspective only the magnitude of the T:R ratio is considered. Our model would output positive and negative T:R ratios (depending on the sign of the T wave), where the magnitude of these outputs would be used for the screening analysis.
This would enable us to examine the T:R ratio for each 10-second segment of each lead from the 24-hour screening.
To demonstrate this, we use this method to predict the T:R ratio for each 10-second segment of each lead of a 3-lead, 24-hour ECG recording.
\begin{figure}
  \centering
  \includegraphics[width=\columnwidth]{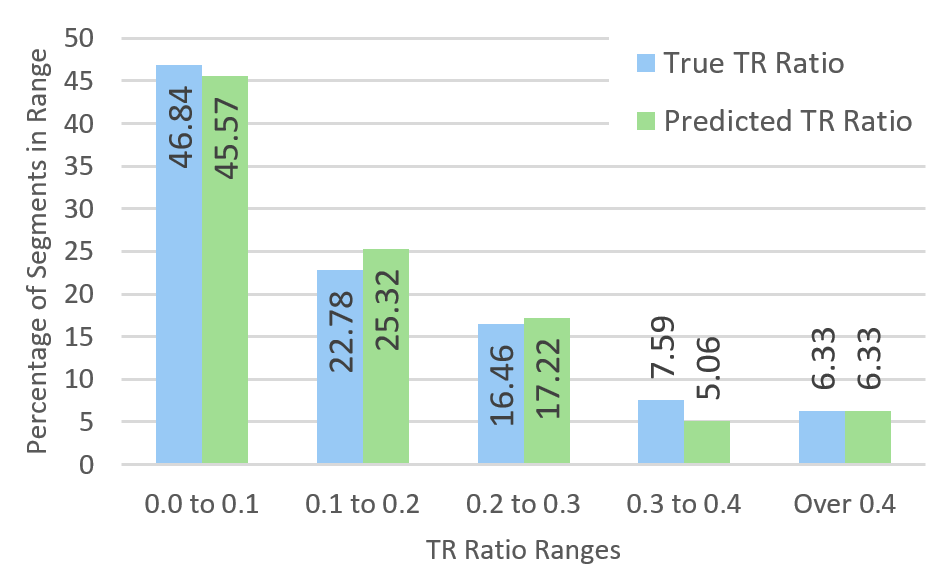}
  \caption{Histogram of true and predicted TR ratios of 10-second ECG segments in the testing set.}\label{preliminary_outputs}
\end{figure}
The primary aim of the screening is to determine if any of the leads are at a low enough risk of TWOS to be used by an S-ICD.
As stated in Section \ref{Introduction}, for a given lead, should a patient have a T:R ratio above 1:3 for a continuous period of at least 20 seconds, that lead would fail the screening.
This means that, for each lead, if our model predicts a T:R ratio of over 0.33 for two or more consecutive 10-second segments, then that lead has failed the screening.

In the event that multiple leads pass the screening, the secondary aim of the screening is to determine which of the leads that pass are at the lowest risk of TWOS.
To examine how the behavior of the T:R ratio differs between each lead, we may wish to plot a histogram of what proportion of the 24-hour screening period the T:R ratio of a particular lead spent in each range of T:R ratios.
To give an example of how our tool is capable of performing this task, Figure \ref{preliminary_outputs} shows the histogram of the T:R ratios predicted by the Complex CNN5 model as well as the histogram of true T:R ratios for PSR images reserved for testing in one of the rounds of cross validation.
As one can see, our model is able to predict the proportion of PSR images in the testing set whose T:R ratio belongs to each range to within 2.5\% of the true value.
Figure \ref{ToolHistograms} contains three histograms showing the proportion of the 24-hour screening that each lead spent with a predicted T:R ratio in each range.
This would enable the cardiologist to assess which of the leads that passed the screening spends the smallest proportion of the 24-hour screening with high values of T:R ratio and, as such, is at the lowest risk of TWOS.

\begin{figure*}[b]
  \centering
  \includegraphics[width=\textwidth]{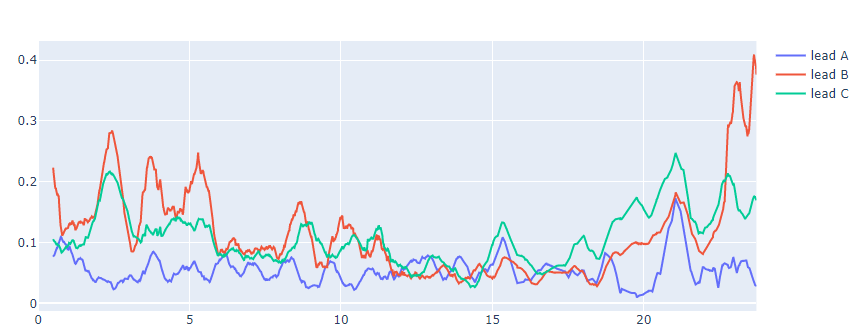}
  \caption{Graph of the variation of the predicted TR ratio over the 24-hour screening preriod for each lead}\label{ToolVariation}
\end{figure*}

Additionally, while less directly applicable to the screening, our model allows for further analysis of the variation of the T:R ratio over the 24-hour screening.
Figure \ref{ToolVariation} shows one tool that our model facilitates. The variation of the T:R ratio is plotted for each lead, over the 24-hour period. For readability, the lines in this graph is smoothed in such a way that each point gives the average T:R ratio for the preceding half hour.
\begin{figure*}[b]
  \centering
  \includegraphics[width=\textwidth]{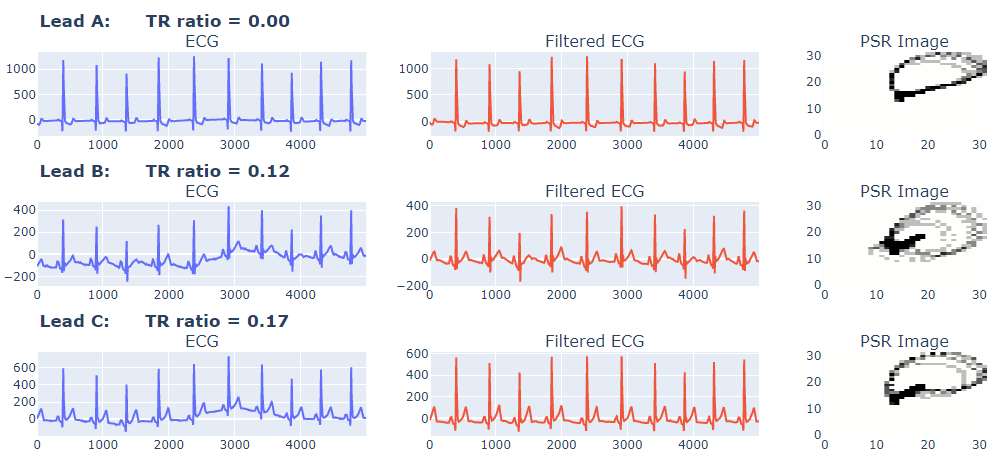}
  \caption{Visualizations of data from a single 10-second segment of the 24 hour ECG recording}\label{ToolSegment}
\end{figure*}
This could enable a cardiologist to detect any period during the 24 hours where the T:R ratio was consistently high and, as such, the patient was at greater risk of TWOS.
Our model could also allow cardiologists to further examine any single 10-second segment from within the 24-hour screening period across all 3 leads and view the ECG signal alongside its predicted T:R ratio.
This is shown in Figure \ref{ToolSegment}, where, for a single 10-second segment, for each lead, the predicted T:R ratio is given alongside a plot of the original ECG signal, a plot of the filtered ECG signal and the PSR image of that filtered ECG signal.

\section{Conclusion}\label{summary}
TWOS is an inherent risk with S-ICDs and can lead to inappropriate shocks.
The current method for determining if a patient is likely to suffer from TWOS is to perform a 10-second screening of the patient's ECG, where the T:R ratio, a major predictor of TWOS, is examined.
Temporal variations in the T:R ratio make this single 10-second screening unreliable.

In this paper, we have developed and tested a convolutional neural network (CNN) based model for predicting T:R ratios from 32x32 pixel PSR images derived from 10-second ECG segments.
As we have shown in Section \ref{Experiments}, this model is capable of predicting T:R ratios for a 10-second ECG segment with a high degree of accuracy.
We have also shown that this model can be integrated into a clinical tool for performing automated screening over much longer periods, such as 24-hours.
This tool is able to much more reliably determine the normal range of a patients T:R ratio than the 10-second screening, as well as being able to give insight into the variation of the T:R ratio over the screening period.
The increased reliability and descriptiveness of this tool can allow cardiologists to better assess the eligibility of patients for S-ICD implantation.

As future work, we have plans to use this tool to analyse the ECG of patient groups suffering from a range of CVDs in order to determine how the behavior of the T:R ratio differs between these groups. Additionally, we hope to determine if the T:R ratio of a patient's ECG can be an indicator of impending cardiac episodes such as VT or VF.

\section*{Acknowledgments}

The work of Anthony J. Dunn is jointly funded by Decision Analysis Services Ltd and EPSRC
through the Studentship with Reference EP/R513325/1. The work of Alain B. Zemkoho is supported
by the EPSRC grant EP/V049038/1 and the Alan Turing Institute under the EPSRC
grant EP/N510129/1.

The feedback provided by Sion Cave (DAS Ltd) on the initial draft of the paper is gratefully acknowledged.

\end{document}